\documentclass{article}

\usepackage{microtype}
\usepackage{graphicx}
\usepackage{subfigure}
\usepackage{booktabs} 

\usepackage{hyperref}

\usepackage[accepted]{icml2024}
\usepackage{mathtools}
\usepackage{amsthm}

\usepackage[utf8]{inputenc} 
\usepackage[T1]{fontenc}    
\usepackage{hyperref}       
\usepackage{url}            
\usepackage{booktabs}       
\usepackage{amsfonts}       
\usepackage{nicefrac}       
\usepackage{microtype}      
\usepackage{amsmath}
\usepackage{amssymb}
\usepackage{multirow}

\usepackage{color,xcolor}
\usepackage{epsfig}
\usepackage{graphicx}

\usepackage{adjustbox}
\usepackage{tabularx}
\usepackage{array}
\usepackage{booktabs} 
\usepackage{colortbl}
\usepackage{float,wrapfig}
\usepackage{hhline}
\usepackage{multirow}

\definecolor{MyDarkBlue}{rgb}{0,0.08,1}
\definecolor{MyDarkGreen}{rgb}{0.02,0.6,0.02}
\definecolor{MyDarkRed}{rgb}{0.8,0.02,0.02}
\definecolor{MyDarkOrange}{rgb}{0.80,0.4,0.04}
\definecolor{MyPurple}{RGB}{111,0,255}
\definecolor{MyRed}{rgb}{1.0,0.0,0.0}
\definecolor{MyGold}{rgb}{0.75,0.6,0.12}
\definecolor{MyDarkgray}{rgb}{0.66, 0.66, 0.66}

\def\prefix#1{\mathbf{y}_{<#1}}
\def\prefixeq#1{\mathbf{y}_{\leq#1}}
\newcommand{\bxy}{\mathbf{x,y}}
\newcommand{\byx}{\mathbf{y|x}}
\newcommand\given[1][]{\:#1\vert\:}

\newcommand{\sidistoryline}[1]{\textcolor{MyGold}{[Storyline: #1]}}

\newcommand{\shanchan}[1]{\textcolor{MyDarkGreen}{[Shanchan: #1]}}
\newcommand{\wenbo}[1]{\textcolor{MyDarkGreen}{[Wenbo: #1]}}
\newcommand{\chenyang}[1]{\textcolor{MyDarkGreen}{[Chenyang: #1]}}
\newcommand{\ziyi}[1]{\textcolor{MyDarkGreen}{[Ziyi: #1]}}

\renewcommand{\sidistoryline}[1]{}
\renewcommand{\shanchan}[1]{}
\renewcommand{\wenbo}[1]{}
\renewcommand{\chenyang}[1]{}
\renewcommand{\ziyi}[1]{}

\begin{document}
\twocolumn[
\icmltitle{DiNADO: Norm-Disentangled Neurally-Decomposed Oracles for Controlling Language Models}



\icmlsetsymbol{equal}{*}

\begin{icmlauthorlist}
\icmlauthor{Sidi Lu}{ucla}
\icmlauthor{Wenbo Zhao}{amazon}
\icmlauthor{Chenyang Tao}{amazon}
\icmlauthor{Arpit Gupta}{amazon}
\icmlauthor{Shanchan Wu}{samsung}
\icmlauthor{Tagyoung Chung}{amazon}
\icmlauthor{Nanyun Peng}{ucla}
\end{icmlauthorlist}

\icmlaffiliation{ucla}{Department of Computer Science, University of California, Los Angeles}
\icmlaffiliation{amazon}{Amazon AGI}
\icmlaffiliation{samsung}{Samsung Research America; Work was done when Shanchan and Sidi were working at Amazon}

\icmlcorrespondingauthor{Sidi Lu}{sidilu@cs.ucla.edu}
\icmlcorrespondingauthor{Nanyun Peng}{violetpeng@cs.ucla.edu}

\icmlkeywords{Machine Learning, ICML}

\vskip 0.3in
]



\printAffiliationsAndNotice{}  

\begin{abstract}
NeurAlly-Decomposed Oracle (NADO) is a powerful approach for controllable generation with large language models. It is designed to avoid catastrophic forgetting while achieving guaranteed convergence to an entropy-maximized closed-form optimal solution with reasonable modeling capacity. 
Despite the success, several challenges arise when apply NADO to a wide range of scenarios. 
Vanilla NADO suffers from gradient vanishing
for low-probability control signals and is highly reliant on a regularization to satisfy the stochastic version of Bellman equation. 
In addition, the vanilla implementation of NADO introduces a few additional transformer layers, suffering from a limited capacity especially compared to other finetune-based model adaptation methods like LoRA. 
In this paper, we propose a improved version of the NADO algorithm, namely DiNADO 
(norm-\textbf{Di}sentangled \textbf{N}eur\textbf{A}lly-\textbf{D}ecomposed \textbf{O}racles), which improves the performance of the NADO algorithm through disentangling the step-wise global norm over the approximated oracle $R$-value for all potential next-tokens, allowing DiNADO to be combined with finetuning methods like LoRA. We discuss in depth how DiNADO achieves better capacity, stability and flexibility with both empirical and theoretical results. Experiments on formality control in machine translation and the lexically constrained generation task CommonGen demonstrates the significance of the improvements. \footnote{Code: \url{https://github.com/PlusLabNLP/DiNADO}}
\end{abstract}
\section{Introduction}
\sidistoryline{Briefly introduce the success of large pretrained models. Introduce the paradigm of finetuning.} Large pretrained generative transformers \citep{radford2019gpt2,brown2020gpt3,raffel2020t5} have achieved remarkable success in a wide range of natural language generation tasks, such as story generation, text summarization, and question answering. Such models benefit from the vast amount of training data to learn powerful distributions that contain rich information about the underlying logic of human languages. 

\sidistoryline{describe the problem of finetuning: 1) as the base model becomes larger, the efficiency of doing so becomes a challenge. some extremely large models are even just provided as a service rather than an open-sourced model. 2) the catastrophic forgetting problem. model finetuned on downstream tasks overfit to the target domain, thus forgetting important knowledge it used to grasp during the pretraining stage. this is even more severe when the model is tested for its reasoning capabilities, typically but not limited to commonsense reasoning.} 
One typical way to adapt such models for specific applications is through fine-tuning. 
However, there are a few problems associated with fine-tuning: 1) The computational efficiency of fine-tuning is highly dependent on the model's number of parameters. Fine-tuning some extremely large models provided as services rather than open-sourced checkpoints can be too expensive for the majority of the community. 2) Fine-tuning on smaller datasets risks causing the catastrophic forgetting problem. A pretrained model can overfit to an under-represented task domain, while forgetting important knowledge it once learned during the pre-training stage. This is particularly a problem when the model is examined for some reasoning capabilities like compositional generalization and/or commonsense reasoning. 

\sidistoryline{briefly introduce prompt tuning: it is a great attempt at resolving the two problems at once. however, with very limited capacity, its flexibility is also pretty much narrowed to specific scenarios. In fact, successful in-context models like ChatGPT usually require much more effort than people originally expect for prompt-based model adaptation. Also, in-context learning requires reading the instructive example thoroughly every time the model is executed} 
Prompt-tuning \citep{dong2022survey} are recent approaches to addressing the challenges associated with fine-tuning large pretrained models. These approaches involve adding a few tokens, which can be discrete natural language tokens or continuous trainable vectors, to the input of a task. Then instead of modifying the parameters of the model, gradient-based optimization is employed to change the embeddings of the added tokens to maximize the probability of the model producing a specific desired output. This allows for the model to adapt to unseen tasks and domains with minimal data, and can also help to avoid the catastrophic forgetting problem associated with traditional fine-tuning methods. However, prompt-tuning has only limited capacity, so it is usually only effective in specific scenarios. 

In-context learning, another popular approach to control/adapt models without needing to update the model parameters, requires reading the prompt/instructive examples \textit{every time} the model is executed. On one hand, this causes computational concerns when a significantly long prompt/instructive example is needed for a complex task. On the other hand, the added tokens or embeddings may not always be able to capture the nuances and complexities of a given task or domain, leading to suboptimal performance.

\sidistoryline{NADO lies between the two: 1) it is implemented with a few (smaller, narrower) transformer layers, which still preserve some model capacity 2) it also does not directly modify the original large pretrained model, which makes it also a good solution to avoid expensive finetuning and catastrophic forgetting. 3) it is proven to adapt the model to the entropy-maximized solution in the ideal case. NADO has shown some initial success in multiple scenarios, including formal machine translation and lexically constrained text generation. However, there are still a few remaining questions when applying such models to solve the controlled generation problem in a compositional manner: 1) when the discrepancy between the controlled distribution and the original distribution is large, what is the best practice to train the NADO layers 2) when can we expect the compositionality of different control signals} 
The NADO algorithm~\cite{tao2022controllable} is a unique approach that lies between fine-tuning and prompt-tuning/in-context learning. It adapts pretrained generative transformers by projecting the base model to the control signal space. To achieve this, NADO is implemented with a smaller transformer model, which controls the base model while preserving necessary model capacity and avoiding direct modification of the base model parameters. 
In the ideal case, NADO adapts the original distribution to the entropy-maximized optimal solution for the target control. It has shown success in various scenarios such as formal machine translation and lexically constrained text generation. 
However, there are still a few open questions when applying such models to solve the controlled generation problems: 1) When the discrepancy between the controlled distribution and the original distribution is large, what is the best practice to train the NADO layers? 2) How can we improve the robustness and efficiency of the NADO module, so that we can reduce additional costs such as the number of samples and additional parameters for training the NADO module? 

\sidistoryline{We address the previous problems in this paper by exploring the best practice of importance sampling. We show that: 1) it is possible to achieve much better training efficiency with importance sampling on a properly selected proxy distribution 2) it is possible to achieve compositional generalization under some necessary yet reasonable assumption} 
In this paper, we address the previous problems related to the NADO algorithm for better training. We propose DiNADO, the improved version of NADO by disentangling the norm of the step-wise value function $R$. DiNADO solves the major issues of vanilla NADO in multiple aspects, leading to better, more stable performance in controllable generation. Our main contributions can be summarized as follows:
\begin{itemize}
    \item We propose an improved parameterization of NADO, namely DiNADO, which generally improves NADO's convergence during both the optional SFT (supervised finetuning) and later stages. We justify the effectiveness empirically and theoretically on the uniqueness of the global parametric optima.
    
    \item We theoretically analyse the inefficiency of the gradient estimation process when using NADO with sparse signals from the control signal function \emph{i.e.} the \emph{oracle}  $C(\mathbf{x}, \mathbf{y})$
    , and demonstrate how to improve the sample/gradient estimation efficiency when training NADO by further exploiting the likelihood predictions from the base model $p$.
    \item We show that with the new formulation, it is natural to combine DiNADO with finetune-based approaches like LoRA, to update the base model $p$ by optimizing the oracle function parameterized by contrasting the altered distribution $q = p_{\phi+\Delta\phi}$ against the base model $p_{\phi}$. This contrastive formulation significantly boosts the model capacity
    of NADO, at the same time allowing for better inference-time performance of the algorithm. 
\end{itemize}

\section{Background and Related Work}
\sidistoryline{Controllable generation for Autoregressive models} \paragraph{Controllable Generation for Autoregressive Models.} There are multiple paradigms to achieve controllable generation with autoregressive models. According to \citet{zhang2022survey}, these paradigms can be classified into three general types: fine-tuning, refact/retraining, and post-processing. Most previous attempts to achieve controllable generation have focused on the first two paradigms, including methods such as CTRL \citep{keskar2019ctrl} and prompt-based learning methods \citep{shin2020autoprompt,letster2021power,li2021prefix}. The post-processing paradigm includes methods such as constrained decoding \citep{anderson2017guided,lu2021neurologic,lu2021astarneurologic} and auxiliary model guided generation \citep{dathathri2020pplm,krause2021gedi,liu2021dexpert,lin2021plug,yang2021fudge}. These methods have shown some success in controlling the base model using signals like lexical constraints, but each of them has its own limitations. Constrained decoding methods fail in directly editing the model distribution, and they may struggle to handle sequence-level control signals that are not trivially factorizable into the token/step level. Auxiliary model guided generation methods have the potential to handle sequence-level, abstract control signals, but they often require additional data or annotation. Moreover, most Auxiliary model guided generation methods don't consider the distribution of the base model, causing distribution discrepancy and degenerated performance in the decoding process.

\sidistoryline{NeurAlly-Decomposed Oracle (NADO)} \paragraph{NeurAlly-Decomposed Oracle (NADO)} NeurAlly-Decomposed Oracle (NADO) \citep{tao2022controllable} is a novel post-processing approach for controllable generation. Given the base distribution $p(\mathbf{x})$ of the large pretrained model NADO controls and the target control signal function $C(\mathbf{x})$, NADO aims to project the base distribution  to the probability space with successful control \emph{i.e.} $C(\mathbf{x}) = 1$. Formally, the target distribution $q(\mathbf{x})$ NADO produces can be written as:

\begin{align}
q(x) = \begin{cases}
\beta p(\mathbf{x}) & \text{if } C(\mathbf{x}) = 1 \nonumber\\
0 & \text{if } C(\mathbf{x}) = 0
\end{cases}
\end{align}
where $\beta$ is the re-normalizing factor that does not need to be calculated explicitly. NADO finds $q(x)$ through learning the step-wise expected satisfaction ratio $R^C(\mathbf{x}_{<t})$, which can be defined as:

\begin{equation}
    R^C(\mathbf{x}_{<t}) = \mathbb{E}_{\mathbf{x}\sim p(\mathbf{x}|\mathbf{x}_{<t})}[C(\mathbf{x})]\nonumber
\end{equation}

where $p(\mathbf{x}|\mathbf{x}_{<t})$ means the distribution of all sequences $\mathbf{x}$ with $\mathbf{x}_{<t}$ as the prefix. 

The vanilla implementation of NADO is naturally limited in both its model capacity and sample efficiency. We argue this is a direct consequence of the entanglement of global norm and the relative strength in the step-wise $R^C(\mathbf{x}_{<t})$. 

\paragraph{The GeLaTo Algorithm} GeLaTo \citep{zhang2023gelato} is an innovative framework aimed at enhancing autoregressive text generation by integrating tractable probabilistic models (TPMs) for imposing lexical constraints. This approach leverages distilled hidden Markov models to effectively guide the generation process in models like GPT-2, ensuring that generated text adheres to specified lexical constraints. Demonstrating superior performance on the CommonGen benchmark, GeLaTo represents a significant advancement in the domain of constrained text generation. It can be formulated as follows:
$$p(x_{t+1} \!\given x_{1:t}, \alpha) \propto {\Pr}_{\text{TPM}}(\alpha \given x_{1:t+1}) \cdot {\Pr}_{\text{LM}}(x_{t+1} \given x_{1:t})$$
Here ${\Pr}_{\text{TPM}}(\alpha \given x_{1:t+1})$ is an HMM-based approximation of the base model.

GeLaTo is capable of achieving perfect control of logic signals, yet requiring the deployment of an HMM-based distribution modifier module. Compared to standard language models, HMM-based models are less scalable, limiting the computational efficiency of GeLaTo.

\sidistoryline{Importance Sampling} \paragraph{Importance Sampling} tackles the problem of calculating an integral by sampling from a different distribution that is easier to sample from, rather than directly from the original distribution, especially when the original distribution is under-sampled. The basic idea behind importance sampling is to reweight the samples generated from a different distribution (known as the \emph{proxy} distribution) so that they are consistent with the target distribution. This reweighting is done using the ratio of the target distribution to the proxy distribution. The formulation of importance sampling can be written as follows:
\begin{equation}
\mathbb{E}_{\mathbf{x}\sim p}[f(\mathbf{x})] = \mathbb{E}_{\mathbf{x}\sim q}[\frac{p(\mathbf{x})}{q(\mathbf{x})}f(\mathbf{x})]\nonumber
\end{equation}

where p(x) is the target distribution, q(x) is the proxy distribution, and f(x) is the function we want to estimate the expected value of. In reinforcement learning, we use importance sampling to estimate the value of a policy under the target distribution by reweighting the returns generated by the proxy policy.

In this work, we use the core idea of the importance sampling and develop upon that to further boost the gradient estimation process of NADO.
 
\section{Methodology}
\sidistoryline{we start off by discussing the challenges to applying the vanilla version of NADO in several scenarios 1) under the previous parameterization of NADO, the parameter of $R$ that corresponds to a particular q(x) is not unique. that is to say, when using q(x) as a likelihood function to warm up the model, the optimization target is ill-defined. to address this, the new parameterization is proposed. 2) with the original random sampling strategy of NADO, it is inefficient to train a controlled model to tackle those C(x) with low satisfaction rate. to address this, we introduce importance sampling, and we hereby discuss how our proxy distribution is constructed. 3) analyze the challenges to directly use NADO in a compositional manner. to address this, we re-adjust the dependencies between different hierarchies of NADOs. with some necessary approximations and assumptions, we can achieve guaranteed compositional generalization.} 
We discuss the challenges of applying the vanilla version of NADO in some tricky scenarios. 
First, when the original distribution $p(x)$ and the target distributions $q(x)$ are far away for each other, NADO usually significantly benefits from an optional warmup step that conducts supervised finetuning (SFT) towards samples from $q(x)$. 
This step is essentially prompt-based fine-tuning like CTRL~\cite{keskar2019ctrl}. 
Under the original parameterization of NADO, during this SFT step, the solution to a particular $q(x)$ is not unique, because the optimization target is ill-defined. 
To address this issue, we first theoretically analyse the major cause of the issue, and then propose a new parameterization of NADO, namely DiNADO (norm-\textbf{Di}sentangled \textbf{N}eur\textbf{A}lly-\textbf{D}ecomposed \textbf{O}racles).

Second, the original random sampling strategy of NADO is inefficient for training a model to tackle control signals with low satisfaction rate. To address this, we introduce importance sampling and discuss how our proxy distribution is constructed.

Finally, we discuss how to further improve the NADO algorithm's capacity and efficiency by combining it with finetune-based adaptation methods like Low-Rank Adaptation (LoRA).
We call the composed algorithm \emph{DiNADO-Merge}. 
DiNADO-Merge allows us to update base model $p$'s parameters by implicitly optimizing the constraint oracle $R$. 
As a result, we obtain the target distribution $q = p_{\phi+\Delta\phi}$ without adding additional parameters. This improves computationally efficient during the inference time as it does not introduce additional parameters while achieve better control over vanilla finetuning with or without LoRA.

\subsection{Notations}
\paragraph{General formulation}Following the notations in the original NADO paper, we use $\mathbf{x}\in \mathcal{X}$ and $\mathbf{y}\in \mathcal{Y}$ to denote the input and generated sequence, respectively. We assume the distributions are defined on the set of sequences of tokens in $\Sigma$.  We denote the $i-$th token in $\mathbf{y}$ as $y_i$ and the sequence prefix from the beginning to the $(i-1)-$th token as $\prefix i$. Thus, for the base auto-regressive language model, the step-wise distribution can be written as
        $p(y_i|\mathbf{x}, \prefix i),$ and the conditional joint distribution as $p(\byx)=\prod_i p(y_i|\mathbf{x}, \prefix i).$ 

\paragraph{Formulation in NADO Modules} We hereby consider the formulation of NADO. The sequence-level oracle can be defined as a boolean function $C:\mathcal{X} \times \mathcal{Y}\rightarrow \{0,1\}.$ We also interchangeably use the notation $C(\mathbf{y})$ or $C(\mathbf{x}, \mathbf{y})$. The resulting step-wise density ratio function can be written as $R^C(\mathbf{x}, \mathbf{y}_{<t})$ or simply $R^C(\mathbf{y}_{<t})$. When we do a one-step enumeration for the next-step likelihood over the vocabulary, we also use the notation $R^C_{\theta}(y_t|\mathbf{y}_{<t}) = \{R^C_{\theta}(\mathbf{y}_{\leq t})\}_{\forall y_t \in \Sigma}$.

\subsection{Normalization Yields Uniqueness of Optima in SFT for NADO-altered likelihood} In the original parameterization of NADO, $R^C_{\phi}(\mathbf{x},\mathbf{y}_{<t})$ reflects the expected value of the decomposed oracle function $C(\mathbf{x},\mathbf{y})$.
To find the optimal $q(y_i|\mathbf{x}, \prefix i)$, it would be natural to assume that $R^C(\mathbf{x},\mathbf{y}_{<t})$ is unique for the optimal a specific $q(y_i|\mathbf{x}, \prefix i)$ induced from $C(\mathbf{x},\mathbf{y})$. However, we have the following lemma to prove that it is not the case:

\noindent \textbf{Lemma 1: The Ambiguious Target to SFT on NADO's Composed Likelihood.} Given the base distribution $p(y_i|\mathbf{x}, \prefix i)$, there are infinite numbers of different unnormalized $R^C(\mathbf{x},\mathbf{y}_{<t})$ (\emph{i.e.} $C(\mathbf{x},\mathbf{y}_{<t})$) that consequent to the same $q(y_i|\mathbf{x}, \prefix i),$.

\textbf{Proof.}~~For an arbitrary real number $0 < \tau < 1$, we can construct a new $C^\tau(\mathbf{x},\mathbf{y}_{<t}) = \tau C(\mathbf{x},\mathbf{y}_{<t})$ and the corresponding $R^{\tau C}(\mathbf{x},\mathbf{y}_{<t})$. We now concern the modified distribution $q^\tau(y_i|\mathbf{x},\prefix{i})$ it induces.

By definition, obviously:
\begin{equation}
\small
    R^{\tau C}(\mathbf{x},\mathbf{y}_{<t}) = \tau R^C(\mathbf{x},\mathbf{y}_{<t})\nonumber
\end{equation}

Since 
\begin{equation}
\small
     q(y_i|\mathbf{x},\prefix{i})=\frac{R^C(\mathbf{x},\prefixeq{i})}{R^C(\mathbf{x},\prefixeq{i-1})}p(y_i|\mathbf{x},\prefix{i}), \nonumber
\end{equation}
and
\begin{align}
     \small
     q^\tau(y_i|\mathbf{x},\prefix{i})\propto&\frac{R^\tau(\mathbf{x},\prefixeq{i})}{R^\tau(\mathbf{x},\prefixeq{i-1})}p(y_i|\mathbf{x},\prefix{i}) \nonumber\\
     =&\frac{\tau R^C(\mathbf{x},\prefixeq{i})}{\tau R^C(\mathbf{x},\prefixeq{i-1})}p(y_i|\mathbf{x},\prefix{i}) \nonumber\\
     =&\frac{R^C(\mathbf{x},\prefixeq{i})}{R^C(\mathbf{x},\prefixeq{i-1})}p(y_i|\mathbf{x},\prefix{i})\nonumber
\end{align}

This implies $q^\tau(y_i|\mathbf{x},\prefix{i}) = q(y_i|\mathbf{x},\prefix{i}).$ Since there are infinite numbers of $\tau$, this successfully disproves the uniqueness of the original unnormalized $R^C(\mathbf{x},\prefixeq{i})$ if we concern a certain $q(y_i|\mathbf{x},\prefix{i})$. While this does not affect the major part of the original NADO algorithm ($R^C(\mathbf{x},\prefixeq{i})$ is still unique given $C$), it leads to an inconsistent objective (and thus sub-optimal performance) during the optional SFT stage (\emph{i.e. warmup}) with NADO. 

We argue that the mitigation of this ambiguity is simple. Given $p$ and $q$, 

\subsection{DiNADO: Disentangling the rescaler from the step-wise oracle factorization value $R$}
We hereby propose a new parameterization of NADO that tackles the problem. In the original NADO algorithm, we directly estimate the decomposition of oracle with a parameterized model $R^C_{\theta}(\mathbf{x},\prefixeq{i})$. In the new parameterization, we try to eliminate the effect of different scaling (\emph{i.e.} $\tau$ in the previous formulation). Without loss of generality, by assuming that $R^C(\mathbf{x},\emptyset) > 0$, where $R^C(\mathbf{x},\emptyset)$ denotes the possibility for the constraints $\mathbf{x}$ to be satisfied when nothing has been generated yet 
(otherwise it would be meaningless to control the base distribution anyway), we instead parameterize a normalized non-negative function $r_{\theta}(\mathbf{x},\prefixeq{i}) \geq 0$ that is in proportion to $R^C(\mathbf{x},\prefixeq{i})$ in each step. Formally:
\begin{align*}
\small
    \forall \mathbf{x}, &\prefix{i}, y_i: \\
    & r_{\theta}(y_i|\mathbf{x},\prefix{i}) = \frac{R^C_{\theta}(\mathbf{x},\prefixeq{i}) }{\beta(\mathbf{x},\prefix{i})}\\
    &s.t. ~~\Vert r_{\theta}(y|\mathbf{x},\prefix{i})\Vert_n = 1.0
\end{align*}

Here $\Vert \cdot \Vert_n$ can be an arbitrary norm. When $n=1$, this is equivalent to $\Vert r_{\theta}(y|\mathbf{x},\prefix{i})\Vert_1 = \sum r_{\theta}(y|\mathbf{x},\prefix{i}) = 1.0$, making $r_{\theta}(\mathbf{x},\prefixeq{i})$ a probability over the vocabulary. 

In all of the following DiNADO variants, we still train the $R^C_\theta$ function (now decomposed into $r^C_\theta$ and $\beta^C_\theta$) following NADO's practice:
\begin{equation*}
    \label{eq:NADOLoss}
    \small
    \begin{aligned}
        L_{CE}(p,R^C_\theta) &=\mathbb{E}_{ \mathbf{y}\sim p(\byx)}L_{CE}(\bxy,R^C_\theta) \\
        &=\sum\nolimits_{\mathbf{y}\in\mathcal{Y}} p(\byx)L_{CE}(\bxy,R^C_\theta) \\
        &=\sum_{i=0}^T CE(R^C_p(\mathbf{x},\prefixeq i), R^C_\theta(\mathbf{x},\prefixeq i))\\
    \end{aligned}
\end{equation*}

Note that in mini-batched training, we adopt the same estimator for the oracle labels $R^C_p(\mathbf{x},\prefixeq i)$ as in NADO with $R^C_p(\mathbf{x},\prefixeq i)=C(\mathbf{x},\mathbf{y})$. It's trivial to prove that this is an unbiased estimation on expectation.

\paragraph{DiNADO-Hard: Towards Regularization-Free Training of NADO} We hereby show that we don't need to parameterize $\beta$ in a sequential manner, but can instead compute it through induction that would eventually result in a more sound formulation of NADO. Without loss of generality, we use L-1 norm in this variant \emph{i.e.} $\sum r_{\theta}(y|\mathbf{x},\prefix{i}) = 1.0$. This makes the output of the NADO module to be a probability over the vocabulary, which is identical to that of a regular language model.

Consider the original regularization used in NADO to ensure the forward consistency condition:
\begin{align}
        \small
            &L_{reg}(\bxy,R^C_\theta)=\nonumber \\
            &f_{KL}\left(\sum_{y_i}R^C_\theta(\mathbf{x},\prefixeq{i})p(y_i|\mathbf{x},\prefix{i}),R^C_\theta(\mathbf{x},\prefixeq{i-1})\right),
            \label{eq:nadov1_reg}
\end{align}

it will only be perfectly satisfied, if and only if:

\paragraph{Equation \ref{eq:forward_consis} (the Forward Consistency Condition)}
\begin{equation}
    \sum_{y_i}R^C_\theta(\mathbf{x},\prefixeq{i})p(y_i|\mathbf{x},\prefix{i}) = R^C_\theta(\mathbf{x},\prefixeq{i-1})
    \label{eq:forward_consis}
\end{equation}

Intuitively, this condition is trying to make sure that the expectation (with the base distribution's probability $p(y_i|\mathbf{x},\prefix{i}$) of $R^C_\theta(\mathbf{x},\prefixeq{i})$ should be \emph{consistent} with the result if we directly take the NADO module's output from the last step $R^C_\theta(\mathbf{x},\prefixeq{i-1})$. From an reinforcement learning's perspective, this condition equation can also be interpreted as the Bellman equation for a stochastic policy agent.

Substituting Equation~\ref{eq:forward_consis} with our rescaler-decomposed parameterization, we have:
\begin{align*}
\small
    &\beta(\mathbf{x},\prefix{i}) \sum_{y_i \in \Sigma}r_{\theta}(\mathbf{x},\prefixeq{i})p(y_i|\mathbf{x},\prefix{i}) \\
    =& \beta(\mathbf{x},\prefix{i-1}) r_{\theta}(\mathbf{x},\prefixeq{i-1})
\end{align*}

Hence we have the following condition for inductively calculating the rescaler $\beta(\mathbf{x},\prefix{i})$: 
\begin{equation}
\small
    \beta(\mathbf{x},\prefix{i}) = \beta(\mathbf{x},\prefix{i-1}) \frac{r_{\theta}(\mathbf{x},\prefixeq{i-1})}{\sum_{y_i}r_{\theta}(\mathbf{x},\prefixeq{i})p(y_i|\mathbf{x},\prefix{i})}
    \label{eq:DiNADO_induction}
\end{equation}

For a unified formulation, it's trivial to prove that $r_{\theta}(\mathbf{x},\emptyset) = 1.0$. We use a classification head to model $\beta_\theta(\mathbf{x},\emptyset)$. 


At inference time, for each step, we can omit $\beta$ and only consider the following modified distribution:

$$q_\theta(y_i|\mathbf{x},\prefix{i}) \propto p(y_i|\mathbf{x},\prefix{i}) r_{\theta}(\mathbf{x},\prefixeq{i})$$

Any likelihood-based SFT on this composed distribution will not have an impact on $\beta(\mathbf{x},\prefix{i-1})$. It's trivial to prove that, given $p(y_i|\mathbf{x},\prefix{i})$, $q_\theta(y_i|\mathbf{x},\prefix{i})$ and $ r_{\theta}(\mathbf{x},\prefixeq{i})$ forms a bi-jection. Note that, without loss of generality, we can assume in practice $\forall \mathbf{x}, \prefix{i}, p(y_i|\mathbf{x},\prefix{i}) > 0, r_{\theta}(\mathbf{x},\prefixeq{i}) > 0, q_{\theta}(\mathbf{x},\prefixeq{i}) > 0$ since they're composed from outputs of neural networks which always predict finite numbers on the log-scale.

\paragraph{DiNADO-Soft: Balancing between Proper Regularization of $R$ and Better Approximation of $C(\mathbf{x}, \mathbf{y})$} While DiNADO-Hard provides a way to completely get rid of the regularization term and potentially improve the controllability, it is possible that the introduced prior of a perfectly-satisfied forward consistency condition can introduce practical difficulties in a better approximation of $\mathbb{E}\Bigl[C(\mathbf{x}, \mathbf{y})\Bigr]$ using $R_{\theta}(\mathbf{x},\prefixeq{i})$.

DiNADO-Soft compromises between the vanilla NADO and DiNADO-Hard. While still adopting the rescaler decomposition parameterization, in DiNADO-Soft we drop the induction in Equation~(\ref{eq:DiNADO_induction}) and directly model the step-wise rescaler $\beta_\theta(\mathbf{x},\prefix{i})$ instead as follows:
\begin{align*}
    R_{\theta}(\mathbf{x},\prefixeq{i}) = \beta_{\theta}(\mathbf{x},\prefix{i}) r_{\theta}(y_i|\mathbf{x},\prefix{i})
\end{align*}
Compared to DiNADO-Hard, by sharing the classifier head among different steps, this will not further introduce model parameters. During training, we still include the regularization (in Equation~(\ref{eq:nadov1_reg})) in vanilla NADO. In our experiments, we show that DiNADO-Soft still improves upon the vanilla version significantly, especially in a faster and better satisfaction of the \emph{forward consistency condition} in Equation~(\ref{eq:forward_consis}).
\begin{figure*}[t]  
  \centering  
    \includegraphics[width=1.8\columnwidth]{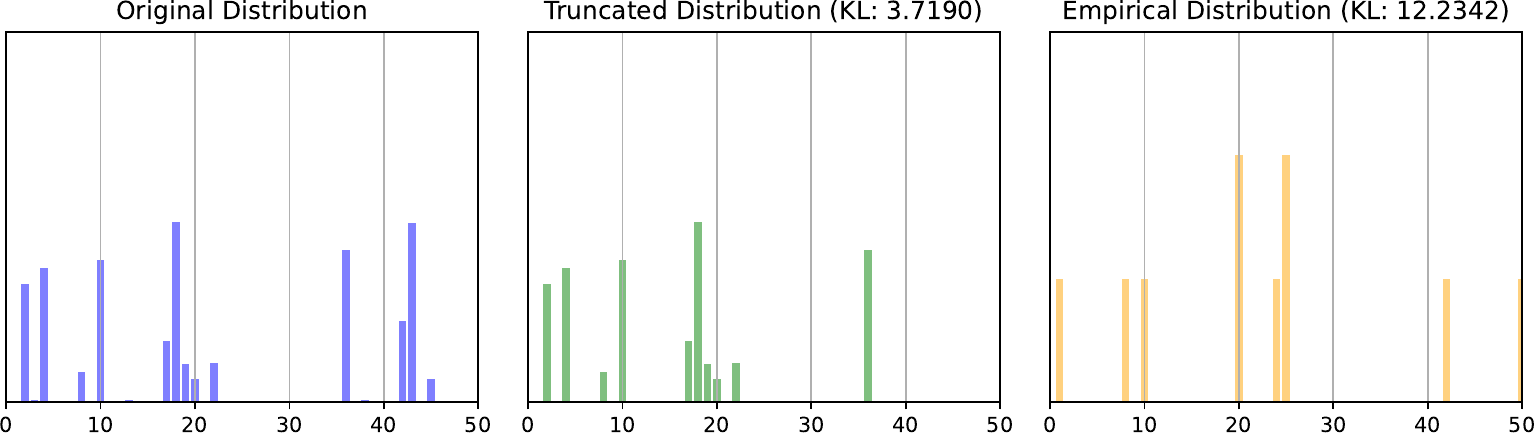} 
  
  \caption{Illustration of the original example distribution, the truncated distribution using the likelihood re-weighting trick and directly approximating the distribution empirically using the same number of random samples.}
  \label{fig:truncate_dist}
\end{figure*}
\paragraph{DiNADO-Merge: Overhead-Free DiNADO through the Comparison of Distributions} We hereby show that with the rescaler-decomposed parameterization, we can utilize finetune-based adaptation methods like LoRA \citep{hu2021lora} to further improve the capacity and scalability of DiNADO. We can achieve this by reversing the cause and effect between the composed distribution $q(y_i|\mathbf{x},\prefix{i})$ and the NADO output $R(\mathbf{x},\prefix{i})$. 

In vanilla NADO, we use the NADO module to directly model $R_\theta(\mathbf{x},\prefix{i})$, and compose the edited distribution $q_\theta(y_i|\mathbf{x},\prefix{i}) \propto p(y_i|\mathbf{x},\prefix{i}) R_\theta(y_i|\mathbf{x},\prefix{i})$. We hereby consider a variant of NADO, where we use LoRA $\Delta\phi=\mathbf{L}^\top\mathbf{U}$ to directly model the edited distribution as $q(y_i|\mathbf{x},\prefix{i}) = p_{\phi+\Delta\phi}(y_i|\mathbf{x},\prefix{i})$, and then train the model with NADO objective through computing $r_{\phi+\Delta\phi}(y_i|\mathbf{x},\prefix{i})$ by contrasting $q(y_i|\mathbf{x},\prefix{i}) = p_{\phi+\Delta\phi}(y_i|\mathbf{x},\prefix{i})$ against the original distribution $p_{\phi}(y_i|\mathbf{x},\prefix{i})$.

For each step, we directly model the rescaler $\beta_\theta(\mathbf{x},\prefix{i})$ and compose $r_{\phi+\Delta\phi}(y_i|\mathbf{x},\prefix{i})$ by L-$\infty$ normalizing $\frac{p_{\phi+\Delta\phi}(y_i|\mathbf{x},\prefix{i})}{p_{\phi}(y_i|\mathbf{x},\prefix{i})}$:
\begin{align}
    &r_{\phi+\Delta\phi}(y_i|\mathbf{x},\prefix{i}) \propto \frac{p_{\phi+\Delta\phi}(y_i|\mathbf{x},\prefix{i})}{p_{\phi} (y_i|\mathbf{x},\prefix{i})} \nonumber \\
    &\max_{y_i} r(y_i|\mathbf{x},\prefix{i}) = 1.0 \nonumber 
\end{align}
\vspace{-1em}
The NADO output can be computed by:

$$R(y_i|\mathbf{x},\prefix{i}) = \beta_\theta(\mathbf{x},\prefix{i}) r_{\phi+\Delta\phi}(y_i|\mathbf{x},\prefix{i})$$

In this case, we can bound $0 \leq \beta_\theta(\mathbf{x},\prefix{i}) \leq 1$, and learn $\beta_\theta(\mathbf{x},\prefix{i})$ as a step-wise binary classification model.

The biggest practical benefit for DiNADO-Merge is that, after training the model to convergence, one have direct access to the modified model $q(\mathbf{y}|\mathbf{x})=p_{\phi+\Delta\phi}(\mathbf{y}|\mathbf{x})$ with no additional inference-time overhead compared to using the original/base model $p_\phi(\mathbf{y}|\mathbf{x})$ alone.

\subsection{Inefficient Gradient Estimation in Vanilla NADO} 
We now concern the variance of gradient w.r.t. $R^C_\theta(\mathbf{x},\prefix{i})$. Suppose we are sampling from the original distribution $p$:
\sidistoryline{Briefly analyse the probability and gradient variance as induced by the probability of getting C(x, y) = 1.0}
\begin{align}
    &\text{Var}(\nabla \mathcal{L}(C; R_\theta; \mathbf{x},\prefixeq{i})) \nonumber\\
    =& \mathbb{E}_{\mathbf{y} \sim p(\mathbf{y}|\mathbf{x},\prefixeq{i})} [ \left(\nabla \mathcal{L}(C; R_\theta; \mathbf{x},\prefixeq{i})\right)^2] \nonumber \\
    -& \mathbb{E}_{\mathbf{y} \sim p(\mathbf{y}|\mathbf{x},\prefixeq{i})}[\nabla \mathcal{L}(C; R_\theta; \mathbf{x},\prefixeq{i})]^2\nonumber\\
    =& \mathbb{E}_{\mathbf{y} \sim p(\mathbf{y}|\mathbf{x},\prefixeq{i})} \left[ \left(\frac{(R_\theta(\mathbf{x},\prefixeq{i}) - C(\mathbf{x},\mathbf{y}))}{R_\theta(\mathbf{x},\prefixeq{i})(R_\theta(\mathbf{x},\prefixeq{i}) - 1)}\right)^2 \right] \nonumber\\
    -& \mathbb{E}_{\mathbf{y} \sim p(\mathbf{y}|\mathbf{x},\prefixeq{i})}[\nabla \mathcal{L}(C; R_\theta; \mathbf{x},\prefixeq{i})]^2 
    \label{eq:grad_analysis}
\end{align}

Without loss of generality, we concern the inefficiency of the gradient estimation when there are still expected updates, \emph{i.e.} 
$$0 < \Vert \mathbb{E}_{\mathbf{y} \sim p(\mathbf{y}|\mathbf{x},\prefixeq{i})}[\nabla \mathcal{L}(C; R_\theta; \mathbf{x},\prefixeq{i})] \Vert < \alpha$$
With a probability $0 \leq \mu \leq 1$, either of the case that our current $R$ disagrees with $C(\mathbf{x}, \mathbf{y})$ is true:
\begin{itemize}
\small
    \item $C(\mathbf{x},\mathbf{y}) = 0$ and $R_\theta(\mathbf{x},\prefixeq{i}) > 1 - \delta$, in this case:
    \begin{align*}
\small
        &\left(\frac{(R_\theta(\mathbf{x},\prefixeq{i}) - C(\mathbf{x},\mathbf{y}))}{R_\theta(\mathbf{x},\prefixeq{i})(R_\theta(\mathbf{x},\prefixeq{i}) - 1)}\right)^2\\
        =&\left(\frac{R_\theta(\mathbf{x},\prefixeq{i})}{R_\theta(\mathbf{x},\prefixeq{i})(R_\theta(\mathbf{x},\prefixeq{i}) - 1)}\right)^2\\
        =&\left(\frac{1}{(R_\theta(\mathbf{x},\prefixeq{i}) - 1)}\right)^2 > \frac{1}{\delta^2}
    \end{align*}
    \item $C(\mathbf{x},\mathbf{y}) = 1$ and $R_\theta(\mathbf{x},\prefixeq{i}) < \delta$, in this case:
    \begin{align*}
    \small
        &\left(\frac{(R_\theta(\mathbf{x},\prefixeq{i}) - C(\mathbf{x},\mathbf{y}))}{R_\theta(\mathbf{x},\prefixeq{i})(R_\theta(\mathbf{x},\prefixeq{i}) - 1)}\right)^2\\
        =&\left(\frac{R_\theta(\mathbf{x},\prefixeq{i}) - 1}{R_\theta(\mathbf{x},\prefixeq{i})(R_\theta(\mathbf{x},\prefixeq{i}) - 1)}\right)^2\\
        =&\left(\frac{1}{R_\theta(\mathbf{x},\prefixeq{i})}\right)^2 > \frac{1}{\delta^2}
    \end{align*}
\end{itemize}

We can then have the following lower bound for Eq.~\ref{eq:grad_analysis}:
\begin{align*}
\small
   \forall 0 \leq \mu, \delta \leq 1:~~ \text{Var}(\nabla \mathcal{L}(C; R_\theta; \mathbf{x},\prefixeq{i})) \geq \frac{\mu}{\delta^2} - \alpha^2 
\end{align*} 

This shows that if $\mu$ is non-neglectable, as $\delta \rightarrow 0$, the variance of the gradient estimation process in vanilla NADO is highly impacted by the sparsity (represented by $\delta$) of the oracle function $C(\mathbf{x},\mathbf{y})$ and can lead to an inefficient training process due to the random detours caused by the noisy gradient from mini-batches.

\paragraph{Tackling Insufficient Presentation of Distribution: Likelihood Re-weighting Importance Sampling.} \sidistoryline{Analyse the keypoints: the significant support of the gradient consists of two parts: those that C(x, y) = 1.0 and those such that C(x, y); Practically better choice: balance the number of real/negative samples, re-normalize with p(x) in-batch.}
We now consider a better strategy for choosing a proxy distribution and perform importance sampling to reduce the variance of gradient. Since in practice, we can only collect a limited number of samples and the empirical distribution derived from the samples do not reflect the original distribution well. See Figure~\ref{fig:truncate_dist}. 
To mitigate this, we collect a \emph{set} (\emph{i.e.} the collection of \textbf{unique} elements) of decoded data as the dimensions we keep to represent the distribution, 
and use the original distribution $p$ to assign a normalized weight for each of the unique basis samples. In practice, this can be achieved by either running random sampling multiple times until having collected a sufficient number of unique samples or simply doing a beam search to approximately select the top-$K$ ($K$ is the sample size limit) of $p$ as the truncation basis. 
It is trivial to prove that this minimizes the $f_{KL}$ between the truncated distribution and original distribution. 

\section{Experiments}
Following the setup of NADO \citep{tao2022controllable} and FUDGE \citep{yang2021fudge}, we evaluate different variants of DiNADO in comparison to the vanilla NADO and other existing controllable decoding methods on the supervised Lexically Constrained Generation (LCG) task using the CommonGen dataset\citep{lin2019comgen} and FormalMT with Fisher and CALLHOME Spanish-English Speech Translation Corpus dataset\citep{post2013improved}. 

\subsection{Dataset Setup}

\paragraph{FormalMT} In both the Spanish source and the original English reference, the sentences are informal and causal. In our evaluation, we follow the setup in NADO \citep{tao2022controllable} to use the rewritten, formal reference \citep{salesky2019fluent} instead. The formality score $C$-function here is approximated by a binary classifier, also adopted from previous works \citep{yang2021fudge,tao2022controllable}.

\paragraph{LCG} CommonGen is designed to evaluate the commonsense reasoning ability of neural text generation models, as well as examining their compositional generalization ability. The training set consists of 32,651 unique key concepts, which serve as constraints, and a total of 67,389 annotated description sequences. Additionally, a validation set containing 993 concepts and 4,018 description sequences is provided. To ensure a comprehensive evaluation, the dataset maintains an open leaderboard for benchmarking different approaches on a withheld test set. However, the maintenance of the test server for this test set has already concluded. As an alternative setup, we use GPT-4~\citep{achiam2023gpt4} to generate the test-set reference. For all other setups, we closely followed previous paper's data configurations to ensure consistency with prior work. 


  

\subsection{Experiment: Better Controllable Generation with DiNADO}
We hereby conduct experiments to study the effectiveness of the proposed approach. We split our experiments into two parts: 1) \textbf{Main results} that utilizes a GPT-2-Large base distribution and compared against existing works on constrained decoding algorithms, evaluated by generation quality (BLEU) and controllability (\emph{Coverage} of the keywords); 2) \emph{Sample efficiency study} to investigate how different designs and objective reweighting tricks help mitigating the sample efficiency issue of NADO.

\subsubsection{Main Result 1 - Input-agnostic Formality Control in Machine Translation}
We first present the primitive results in comparison with existing algorithms on the formality control problem in machine translation.
\begin{table}[h]
  \centering
  \small
  \caption{Performance of different ways for adapting and controlling the language model on FormalMT. Results with $*$ are reported from the original paper. } 
  
  \setlength{\tabcolsep}{1mm}
  \begin{tabular}{lcc}
    \toprule
    \toprule
    Model & Formal BLEU & Formality \\
    \toprule
    GPT-2 Large \\
    \midrule
    Direct Finetune & 16.84 & 0.44\\
    \midrule
    NADO & 20.76 & 0.53 \\ 
    \midrule
    DiNADO-Soft-GPT-2 Base & 21.06 & 0.59\\
    DiNADO-Merge-LoRA & 21.37 & 0.59 \\
    DiNADO-Merge-fullparameter & \textbf{21.58} & \textbf{0.60} \\
    \toprule
    MarianMT \\
    \midrule
    Direct Finetune & 16.98$^*$ & 0.45$^*$ \\
    \midrule
    FUDGE & 17.96$^*$ & 0.51$^*$ \\
    NADO & 21.04$^*$ & 0.53$^*$ \\
    \bottomrule
    \bottomrule
  \end{tabular}%
  \label{tab:main_results_mt}
\end{table}
DiNADO-Merge with full parameter finetuning achieves the best performance among all models, followed by DiNADO-Merge with LoRA.
\subsubsection{Main Result 2 - Input-aware Controlled Generation for CommonGen}

We now present the second part of main results in comparison with existing algorithms. We focuses on the supervised setting, and report the results with either the self-reported ones or our re-evaluation with the synthesized references.
\begin{table}[h]
  \centering
  \caption{Performance of different ways for adapting and controlling the language model on CommonGen. Results with * are evaluated with the GPT-synthesized pseudo test references.}
  
  \setlength{\tabcolsep}{1mm}
  \resizebox{0.99\columnwidth}{!}{%
  \begin{tabular}{lcccc}
    \toprule
    \toprule
    Model & \multicolumn{2}{c}{BLEU} & \multicolumn{2}{c}{Coverage} \\
    & (dev) & (test) & (dev) & (test) \\ 
    \toprule
    \multicolumn{3}{l}{GPT-2 Large}\\
    \midrule
    Direct Finetune & 27.6 & 24.1* & 87.4\% & 87.2\% \\
    NeuroLogic \citep{lu2021neurologic} & - & 26.7 & - & 96.7\% \\
    A*esque \citep{lu2021astarneurologic} & - & 28.2 & - & 97.8\% \\
    \midrule
    GeLaTo \citep{zhang2023gelato} & 34.0 & 34.1 & 100\% & 100\% \\
    - (w/o rerank) & 32.5 & 32.9* & 100\% & 100\% \\
     \midrule
    NADO-Adaptation Layers & 30.3 & 30.1* & 97.1\% & 96.2\% \\
    NADO-GPT-2 Base & 30.8 & 30.7* & 97.6\% & 96.8\% \\
    \midrule
    DiNADO-Hard-Adaptation Layers & 28.3 & 28.4* & 97.5\% & 97.6\% \\
    DiNADO-Hard-GPT-2 Base & 28.9 & 29.0* & 97.7\% & 97.9\% \\
    DiNADO-Soft-Adaptation Layers & 29.9 & 30.0* & 97.8\% & 97.5\% \\
    DiNADO-Soft-GPT-2 Base & 31.6 & 31.4* & 98.6\% & 97.9\% \\
    DiNADO-Merge & \textbf{34.3} & \textbf{34.2}* & 98.5\% & 98.9\% \\
    - (w/o rerank) & 32.3 & 32.9 & 98.4\% & 98.9\% \\
    
    \bottomrule
    \bottomrule
  \end{tabular}%
  }
  \label{tab:main_results_lcg}
\end{table}

As is shown in Table~\ref{tab:main_results_lcg}. DiNADO-Merge achieves the state-of-the-art among all constrained decoding algorithms.

\subsubsection{Discussion}
\begin{figure}[h!] 
  \centering
  \subfigure[Major (Class-entropy) Loss]{\includegraphics[width=0.48\columnwidth]{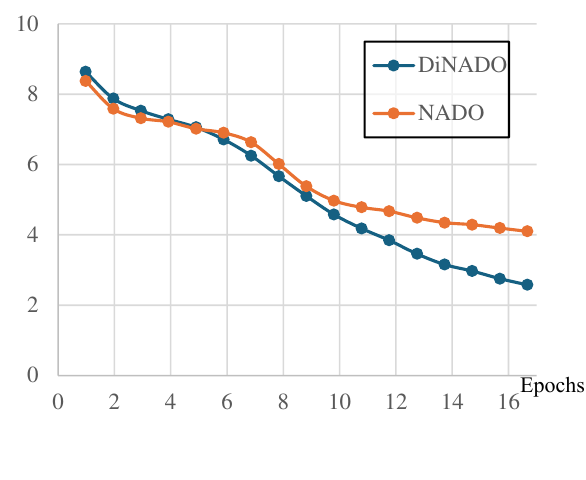}} \label{fig:compare_stability_class}
    \subfigure[Forward Consistency Regularization Loss]{\includegraphics[width=0.48\columnwidth]{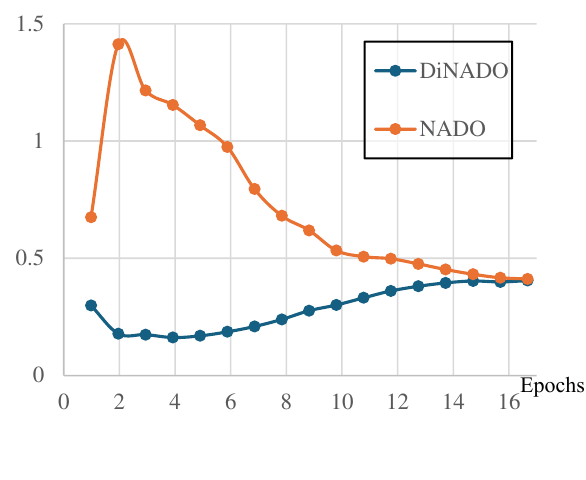}}\label{fig:compare_stability_reg}
   \caption{(a) The original NADO and DiNADO converge with similar dynamics at early stages in terms of the major part of loss, but DiNADO converges to a better local optima. (b) DiNADO's norm disentanglement significantly helps to stabilize the regularization term.}  
   \label{fig:compare_stability}
\vspace{-15pt} 
\end{figure}  

\paragraph{DiNADO-Hard versus DiNADO-Soft} While the perfect satisfaction of the \emph{forward consistency condition} is theoretically appealing, we find that in practice this hinders the effective learning of the oracle signal $C(\mathbf{x}, \mathbf{y})$, resulting in an inferior performance of DiNADO-Hard generally. In particular, with an inaccurate estimation of the initial $\beta_\theta(\mathbf{x},\emptyset)$, with the false estimation being too high, in intermediate steps it could be possible that $R^C_\theta(\mathbf{x},\prefixeq{i}) > 1$ which violates the definition of $R^C_\theta(\mathbf{x},\prefixeq{i}) > 1$. We argue that this further negatively impact the performance of the composed distribution.

\paragraph{Adaptation Layer versus Smaller Model versus DiNADO-Merge} The larger capacity generally grant the NADO module with more flexibility and the composed model better controllability. This observation is generally consistent with the conclusions from our discussion in Section~\ref{sec:likelihood-SFT}. DiNADO-Merge in particular shows more generalizable performance, which we argue can be a direct result that it manage to handle the controllability in the same parameter space as the pretrained base model.

\paragraph{Discussion: Training Stability Comparison} We further conduct a study on the training stability of DiNADO against NADO through tracing the training loss of them. See Figure~\ref{fig:compare_stability}. In Figure~\ref{fig:compare_stability_class}, we find that the original NADO and DiNADO converge with similar dynamics at early stages in terms of the major part of the loss which are most associated with the learning of the $R$ function. However, DiNADO is capable of more sufficiently approximating $R$, especially after around 10 epochs of training. In Figure~\ref{fig:compare_stability_reg}, we observe that the regularization term in vanilla NADO first increases due to divergence from the original distribution, and then converges slowly until below 0.5. According to previous study in NADO, the divergence of the regularization term would harm how the improvement in $R$ actually helps the performance of the modified distribution $q$. In contrast, the norm-disentanglement in DiNADO helps it to always control the regularization term under 0.5, making most updates in $R$ always helpful towards the improvement of the composed distribution $q$.

\begin{table}[t]
  \centering
  \small
  \caption{Post-SFT performance of different parameterization for distribution post-editing on CommonGen. The number in parenthesis indicates the unremovable extra parameter number for the $R$ module. The behavior of \emph{Unnormalized} corresponds to the behavior of NADOv1 during its optional SFT phase, whereas \emph{Normalized} corresponds to DiNADO(-Soft/Hard).}
  \resizebox{0.99\columnwidth}{!}{%
  \begin{tabular}{lcc}
    \toprule
    \toprule
    Model & BLEU (dev) & Coverage \\
    \toprule
    \multicolumn{3}{l}{GPT-2 Large}\\
    \midrule
    Direct Finetune (GPT-2 Large) & 27.61 &  87.4\% \\
    Direct Finetune (GPT-2 Base) & 17.36 &  84.1\% \\
    \midrule
    Unnormalized-Adaptation Layers (119M) & 13.98 &  86.8\% \\
    Unnormalized-GPT-2 Base (117M) & 15.12 &  87.1\% \\
    Normalized-Adaptation Layers (119M) & 16.36 & 87.2\% \\
    Normalized-GPT-2 Base (117M) & 17.15 & 87.3\% \\
    Normalized-Merge (0M) & 28.91 & 88.5\% \\
    
    \toprule
    \multicolumn{3}{l}{T5 Large}\\
    \midrule
    Direct Finetune (T5 Large) & 30.33 & 93.3\% \\
    Direct Finetune (T5 Small) & 22.31 & 91.5\% \\
    \midrule
    Unnormalized-Adaptation Layers (92M) & 16.71 & 87.1\% \\
    Unnormalized-T5 Small (80M) & 18.22 &  86.9\% \\
    Normalized-Adaptation Layers (92M) & 19.87 & 86.2\% \\
    Normalized-T5 Small (80M) & 20.49 & 87.7\% \\
    Normalized-Merge (0M) & 30.73 & 92.9\% \\
    \toprule
    \multicolumn{3}{l}{Instructed Flan-T5}\\
    \midrule
    Direct Instruct (Flan-T5 Large) & 34.53 & 95.8\% \\
    Direct Instruct (Flan-T5 Small) & 18.80 & 82.9\% \\
    \midrule
    Unnormalized-Adaptation Layers (92M) & 32.89 & 92.8\% \\
    Unnormalized-T5 Small (80M) & 33.50 & 96.3\% \\
    Normalized-Adaptation Layers (92M) & 31.67 & 93.4\% \\
    Normalized-T5 Small (80M) & 33.84 & 96.2\% \\
    Normalized-Merge (0M) & 34.67 & 96.5\% \\
    \bottomrule
    \bottomrule
  \end{tabular}%
  }
  \label{tab:post_warmup_results}
\end{table}
\subsubsection{Ablation Study: Experiment on Likelihood-based SFT (Warm-up)} 
\label{sec:likelihood-SFT}
We first conduct an experiment on the CommonGen dataset to study how much the proposed new parameterization of NADO alleviates the known issues during the likelihood-based warmup process. We consider a harder case than the original warmup process in the original NADO paper. Instead of training the base distribution to be an unconditional description model, we now concern directly using NADO to adapt pretrained language models to handle the CommonGen task without any task-specific finetuning of the base distribution. We view this experiment as an examination of the extra model capacity introduced by different ways of model controlling.

We discuss and compare the post-warm-up performance under two different cases: 1) finetuning pretrained models without instruction tuning (e.g. GPT-2 \citep{radford2019gpt2}) and unstructured prompting; 2) instructing and finetuning pretrained model with instruction tuning (e.g. Flan-T5~\citep{chung2022scaling}). See Table~\ref{tab:post_warmup_results}.

In addition, we compare different ways to model the probability mask $R^C_{\theta}(\mathbf{x},\prefixeq{i})$. In addition to the original way of adaptation layers described in NADO~\citep{tao2022controllable}, we also examine the case where we use a fully independent yet smaller model to serve as $R^C_{\theta}(\mathbf{x},\prefixeq{i})$.

\paragraph{Controlling the Base Model: Adaptation Layers versus a Smaller Model} We find that using a smaller model in general performs better than the original way of using adaptation layers (of similar parameter numbers) for both NADO and DiNADO. Note that, since the objective in the NADO algorithm is very different than that from the original language model initialization, for language models that use tied input-output embeddings, we need to untie them for effectively adapting the model as the NADO module. Notably, experiment results using Flan-T5 show that, for cases where the base distribution is good enough, the introduction of adaptation layers can cause the composed distribution to be even worse than simply direct sampling from the base distribution.

\paragraph{Unnormalized versus Normalized: Improved Warm-up with Algorithmic Consistency Mitigation} We find that the mitigation of algorithmic consistency of the original composed likelihood function in DiNADO generally improved the post-SFT performance. When the base model has a huge distribution gap from the target domain (as shown in the case of finetuning GPT-2\citep{radford2019gpt2} and T5 \citep{raffel2020t5} models), this improvement is significant. As a result, the composed distribution after SFT is performing similarly well as directly an SFT model with similar additional parameter capacity. By further combining with finetuning methods like LoRA\citep{hu2021lora} as the DiNADO-Merge algorithm, the post-SFT performance is on par or even better than full model finetuning. This is expected, as the behavior of DiNADO-Merge during SFT stage is identical to vanilla LoRA-finetuning.

\begin{table}[h]
  \centering
  \caption{Performance of different ways for adapting and controlling the language model on CommonGen. Results with * are evaluated with the GPT-synthesized pseudo test references.}
  
  \setlength{\tabcolsep}{1mm}
  \resizebox{0.99\columnwidth}{!}{%
  \begin{tabular}{lcccc}
    \toprule
    \toprule
    Model & \multicolumn{2}{c}{BLEU} & \multicolumn{2}{c}{Coverage} \\
    & (dev) & (test) & (dev) & (test) \\ 
    \toprule
    \multicolumn{5}{l}{GPT-2 Large}\\
     \midrule
    NADO-GPT-2 Base (N=2) & 24.0 & 22.6* & 85.1\% & 84.2\% \\
    NADO-GPT-2 Base (N=4) & 25.1 & 23.4* & 87.4\% & 87.0\% \\
    NADO-GPT-2 Base (N=8) & 27.4 & 26.6* & 91.3\% & 88.7\% \\
    NADO-GPT-2 Base (N=16) & 29.1 & 28.6* & 93.8\% & 92.9\% \\
    NADO-GPT-2 Base (N=32) & 30.8 & 30.7* & 97.6\% & 96.8\% \\
    NADO-GPT-2 Base (N=64) & 30.8 & 30.6* & 97.8\% & 96.9\% \\
    \toprule
    \multicolumn{5}{l}{w/o Likelihood-based Reweighting}\\
    \midrule
    DiNADO-Soft-GPT-2 Base (N=2) & 24.3 & 23.7* & 86.1\% & 85.7\% \\
    DiNADO-Soft-GPT-2 Base (N=4) & 26.6 & 25.1* & 92.7\% & 91.0\% \\
    DiNADO-Soft-GPT-2 Base (N=8) & 28.8 & 28.0* & 94.5\% & 93.9\% \\
    DiNADO-Soft-GPT-2 Base (N=16) & 29.4 & 29.0* & 96.1\% & 95.3\% \\
    DiNADO-Soft-GPT-2 Base (N=32) & 30.9 & 30.8* & 97.1\% & 97.0\% \\
    DiNADO-Soft-GPT-2 Base (N=64) & 30.9 & 30.8* & 97.0\% & 97.3\% \\
    \toprule
    \multicolumn{5}{l}{w/ Likelihood-based Reweighting}\\
    \midrule
    DiNADO-Soft-GPT-2 Base (N=2) & 25.3 & 24.7* & 88.6\% & 87.3\% \\
    DiNADO-Soft-GPT-2 Base (N=4) & 28.6 & 29.1* & 96.7\% & 96.1\% \\
    DiNADO-Soft-GPT-2 Base (N=8) & 30.6 & 30.5* & 96.6\% & 96.9\% \\
    DiNADO-Soft-GPT-2 Base (N=16) & 30.9 & 31.0* & 97.5\% & 97.3\% \\
    DiNADO-Soft-GPT-2 Base (N=32) & 31.3 & 31.2* & 97.9\% & 97.6\% \\
    DiNADO-Soft-GPT-2 Base (N=64) & 31.6 & 31.4* & 98.6\% & 97.9\% \\
    
    \bottomrule
    \bottomrule
  \end{tabular}%
  }
  \label{tab:sample_eff}
\end{table}

\subsubsection{Sample Efficiency Study}
In the original setup, one need to collect 32 independent samples for each unique input to train the NADO module. However, empirically this can still be too expensive, as we would like to investigate how the previously proposed improvements contribute to better sample efficiency of the algorithm. We take two variants of the models as the representative of NADOv1 and DiNADO respectively: NADOv1-GPT-2 Base and DiNADO-Soft-GPT-2 Base. We start from collecting (N=2) samples, and gradually double the sample size until 64 samples per unique input. The results are shown as in~\ref{tab:sample_eff}:

\textbf{Discussion} A sufficient number of samples per unique input is crucial towards the success of NADOv1. The performance of NADOv1 saturates near (N=32), this validates that setting (N=32) is a reasonable choice. Compared to NADOv1, simply improving the algorithm with parameterization as DiNADO-Soft helps in improving the sample efficiency, yet its performance is still impacted by the number of samples to an observable degree. Generally, DiNADO without the likelihood-based reweighting shows a similar dynamics as NADOv1 with a larger sample size. With the likelihood-based reweighting, we can effectively reduce the sample size to as small as (N=4) without significantly sacrificing the resulting performance, although a larger sample size can still boost the performance slightly.

\section{Conclusion}
In this paper, we discuss the existing algorithmic problems of NeurAlly-Decomposed Oracle (NADO), a trainable controllable generation decoding algorthm for language models. Our discussion focuses on theoretically analysing the flawful designs of the vanilla implementation of NADO and proposing respective mitigations, resulting in the improved version of NADO, namely DiNADO as in norm-\textbf{Di}sentangled \textbf{N}eur\textbf{A}lly-\textbf{D}ecomposed \textbf{O}racles. In addition, the new implementation of DiNADO allows it to be naturally combined with finetuning methods like LoRA, resulting in a capacity-rich version of NADO. Experiments on MarianMT and CommonGen justify the significance of these algorithmic improvements.

\section*{Impact Statement}
The goal of this paper is to analyze and mitigate the existing problems in the constrained decoding algorithm NADOv1. For the broader impact, we argue that DiNADO provides an alternative way to achieve the alignement and \emph{super-alignment} of language models. In particular, we show that the feedback from either the smaller language models or a comparison between an LLM and its previous versions could utilize NADO as a more efficient way to control and guide LLMs to better correlate with human preferences with solid theoretical guarantee.

\section*{Acknowledgement}

This research is supported by Amazon AGI and the Amazon internship program. We would like to deeply thank all anonymous reviewers for their insightful feedbacks, as well as our teammates from Amazon including (but not limited to) Anjali Narayan-Chen, Jiun-Yu Kao, Vivek Subramanian and Haw-Shiuan Chang. We would also like to express our deep appreciation to members from UCLA, including (but not limited to) Tao Meng, Zi-Yi Dou, Honghua Zhang, Te-Lin Wu, Po-Nien Kung, Meihua Dang and Prof. Guy Van den Broeck, for their valuable feedback and initial discussions. 


\newpage

\bibliography{nadopp}
\bibliographystyle{acl_natbib}



\end{document}